\documentclass[10pt,twocolumn,letterpaper]{article}
\usepackage[margin=0.72in,columnsep=0.25in]{geometry}
\usepackage[T1]{fontenc}
\usepackage[utf8]{inputenc}
\usepackage{lmodern}
\usepackage{microtype}
\usepackage{amsmath}
\usepackage{booktabs}
\usepackage{array}
\usepackage{tabularx}
\usepackage{xcolor}
\usepackage{graphicx}
\usepackage{placeins}
\usepackage{xurl}
\usepackage{enumitem}
\usepackage{caption}
\usepackage{titlesec}
\usepackage{balance}
\titleformat{\section}{\large\bfseries\raggedright}{\thesection}{0.8em}{}
\titleformat{\subsection}{\normalsize\bfseries\raggedright}{\thesubsection}{0.8em}{}
\usepackage{fancyhdr}
\usepackage[hidelinks]{hyperref}
\hypersetup{
  pdftitle={Signals: Continuous Diagnostics of AI Agents},
  pdfauthor={Amey Rambatla, Vatsal Goel, Sagar Gaur, Siddhant Loya, Anurag Mishra, Asif Mohammad},
  pdfsubject={A diagnostic taxonomy and continuous diagnostics for AI agents}
}
\definecolor{signalblue}{HTML}{163B5C}
\definecolor{lightblue}{HTML}{EEF4F8}
\definecolor{outlinegray}{HTML}{637180}
\setlist[itemize]{leftmargin=1.2em,topsep=3pt,itemsep=2pt,parsep=0pt}
\setlist[enumerate]{leftmargin=1.45em,topsep=3pt,itemsep=2pt,parsep=0pt}
\newcommand{\code}[1]{\texttt{#1}}
\title{\vspace{-1.8em}\LARGE\bfseries Signals: Continuous Diagnostics of AI Agents}
\author{
\begin{tabular*}{0.70\textwidth}{@{}l@{\extracolsep{\fill}}c@{\extracolsep{\fill}}r@{}}
\normalsize Amey Rambatla & \normalsize Vatsal Goel & \normalsize Sagar Gaur\\
\normalsize Siddhant Loya & \normalsize Anurag Mishra & \normalsize Asif Mohammad
\end{tabular*}\\[1.35em]
\small Cisco\\[-0.05em]
\small\texttt{\{ameyr,vatsalgl,gaur,siddhant,anumish,khasmoha\}@cisco.com}
}
\date{}

\begin{document}
\raggedbottom
\twocolumn[
\begin{@twocolumnfalse}
\maketitle
\vspace{-1.0em}
\begin{abstract}

AI agents can make incorrect decisions or waste resources even when their tools and infrastructure work normally. Their nondeterministic behavior makes these problems difficult to reproduce and diagnose. We developed Signals to identify such problems from agentic traces and suggest \textbf{concrete remediations}. We built the method around a taxonomy of \textbf{45 problem types across eight families}, with definitions and evidence requirements for each type. The method distinguishes realized failures, risk exposures, and inefficiencies, groups related findings into engineering issues called Signals, and assesses their severity from the supporting evidence. Each Signal includes actions a developer can take and checks for evaluating the change. Further executions can add evidence to an existing Signal or reveal a new problem. We evaluated the method on public agent executions and achieved \textbf{91.43\% F1} when matching Signals to reference issues, \textbf{about 18 percentage points higher} than a one-pass LLM control that used the same model and taxonomy.

\end{abstract}
\vspace{0.3em}
\vspace{1.2em}
\end{@twocolumnfalse}
]

\section{Introduction}\label{sec:intro}
An AI agent can use the wrong tool, overlook required information, repeat unnecessary work, or claim that a task succeeded when it did not. These failures need not produce an exception or a failed service request. The software can run as written while the agent fails to do what the user intended.

Diagnosing this behavior is difficult because an agent's decisions are not fixed by its code alone. It may interpret information differently or choose a different sequence of actions across runs of the same task. Benchmarks such as $\tau$-bench studied reliability across repeated executions~\cite{r1}. One successful run therefore provides only one observation, and a remediation needs to be checked beyond the execution that motivated it.

Agentic traces give developers a record of model interactions, retrieved information, tool calls, and intermediate results. They do not, by themselves, explain whether the agent made the right decision. A successful tool call can perform the wrong action, and several valid calls can form an unnecessary loop. To diagnose either case, a developer needs to compare the execution with the task, the available information, and the requirements governing the actions.

We developed Signals to perform this analysis continuously and connect the findings to work a developer can undertake. We started with a taxonomy of 45 problem types across eight families, drawing on research into agent failures, retrieval, tool use, and safety~\cite{r1,r2,r3,r4,r5}. The definitions distinguish problems that can lead to similar outcomes but require different fixes. Invalid tool arguments, an invalid tool result, and an agent ignoring a valid result are three such cases.

We then used the taxonomy to connect individual findings to related issues, supporting evidence, severity, and remediation suggestions. The distinction between a finding and an issue matters: repeated instances of the same integration problem may need one fix, while two problems with the same taxonomy label may need different fixes. New executions can strengthen an existing finding, establish a separate issue, or provide evidence about behavior after a change. The taxonomy supplies a consistent vocabulary even as the models and analysis methods used to apply it change.

Our contributions were:
\begin{itemize}
\item \textbf{A taxonomy of agent problems.} We defined 45 problem types across eight families, with evidence requirements and boundaries between related types.
\item \textbf{A continuous diagnostic method.} We connected findings from agent executions to Signals that describe the issues developers need to address.
\item \textbf{Evidence-based severity.} We defined separate factors for realized failures, risk exposures, and inefficiencies, together with a fixed calculation and Signal-level aggregation.
\item \textbf{Concrete remediation suggestions.} We connected each diagnosis to corrective actions and checks, using the available evidence to determine how specific the recommendation could be.
\end{itemize}

\section{Related Work}\label{sec:related}
We drew on agent benchmarks and failure-diagnosis research to understand which problems a diagnostic system needed to recognize. $\tau$-bench evaluated interactions among an agent, a user, tools, and domain policies; its $\mathrm{pass}^{k}$ metric measures success across repeated executions of the same task~\cite{r1}. MAST studied multi-agent failures through annotated execution histories and a failure taxonomy~\cite{r2}. AgentRx used a cross-domain taxonomy, constraints, and supporting evidence to diagnose failures within an execution~\cite{r3}. These studies examined the decisions and interactions leading to an outcome, rather than relying only on whether the task succeeded. SWE-bench evaluated whether language models resolved software issues and provided recorded coding-agent trajectories, adding a repository-based setting for studying agent behavior~\cite{r6}.

Other work examined specific requirements that an agent's output or actions must satisfy. RAGTruth used human annotations to study statements that were unsupported by, or contradicted, retrieved information~\cite{r4}. AgentDojo evaluated prompt injection attacks and defenses under explicit task and attack conditions~\cite{r5}. These settings distinguish an incorrect answer from an attempted attack or a successful defense. The execution record must be interpreted against the task and the requirement that applied to the action.

We used these distinctions when defining the problems in our taxonomy. MAST and AgentRx informed our treatment of coordination, verification, and tool-result interpretation. RAGTruth informed the boundary around unsupported claims, while $\tau$-bench and AgentDojo supplied tool-use, policy, and trust-boundary examples. We then connected the problem definitions to the evidence needed for a diagnosis and to concrete remediation actions. Evaluating false findings requires independently reviewed executions with no supported problem; a benchmark's unsuccessful-task label alone cannot supply that distinction.

\section{Method}\label{sec:method}
\begin{figure*}[t]
\centering
\includegraphics[width=\textwidth]{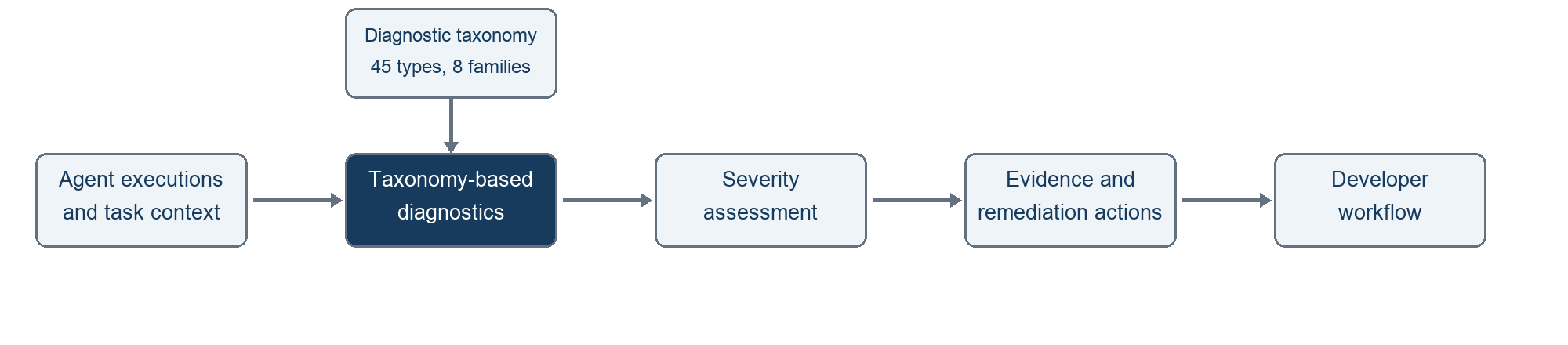}
\caption{The Signals diagnostic workflow. We used the taxonomy to diagnose problems in execution records, assess their severity, and suggest remediation actions. Further executions can add evidence to existing Signals or reveal new issues.}
\label{fig:workflow}
\end{figure*}

We built Signals to turn agent executions into diagnoses that developers could investigate and act on. We first used the taxonomy in Section~\ref{sec:taxonomy} to identify problem occurrences and collect their execution evidence. We then grouped related occurrences into Signals, assessed their severity, and suggested remediations. Figure~\ref{fig:workflow} shows this workflow.

We distinguished three units throughout this process. A \emph{session} is one recorded agent execution. An \emph{occurrence} is an individual problem instance. A \emph{Signal} groups related occurrences into an engineering issue, with an explanation, supporting evidence, severity, and concrete remediation actions. A session can contain several problems, while the same issue can recur across sessions.

\subsection{Finding problems in an execution}
We examined each session against its task request and any supplied instructions or tool contracts. The available record could include model inputs and outputs, retrieved information, tool arguments and responses, and the final result. We interpreted events in that sequence: a rejected call followed by recovery is different from the same rejection followed by task failure. Repeated retrieval can supply needed evidence or repeat work that adds no benefit.

For each finding, the diagnosis identifies the affected component, ranks candidate problem types, explains what happened and where, and links to the records that support it. An error violates an applicable requirement or prevents the task from being completed correctly. An inefficiency adds avoidable cost or latency without a corresponding quality, safety, or reliability benefit. Neither successful service responses nor repeated steps establish either diagnosis on their own. The task, surrounding execution, and applicable requirements determine whether a problem occurred.

We also separated the problem type from its observed outcome. A \emph{realized defect} has an observed adverse consequence, incorrect result, or failure. A \emph{risk exposure} has evidence of a reachable harmful path whose harm was not observed. An \emph{inefficiency} leaves the outcome acceptable but adds work without a supported benefit. The same type can support different outcomes when the evidence warrants them. Missing telemetry, attack-like input, or a long execution is not sufficient evidence of a failure or waste.

\subsection{Generating and updating Signals}
We grouped related findings around the issue a developer would need to fix. Repeated instances of an argument-construction error, for example, can support one Signal with several cited occurrences. A different problem requiring a separate intervention remains a separate Signal. Sharing a taxonomy type alone does not establish a common root cause or remediation.

Further executions can add evidence to an existing issue or establish a new one. When warranted, Signals updates the affected sessions, evidence, severity, and proposed actions. After a remediation, new executions can also provide evidence about recurrence and preserved behavior. One corrected execution, or no findings during an irrelevant period, does not establish that a fix worked. A change still needs independent checks on affected cases and cases that previously succeeded.

\subsection{Assessing severity}\label{sec:severity}
We treated severity separately from problem type and diagnostic confidence. It describes the consequence, reachable exposure, or avoidable waste supported by the evidence. Teams may use that assessment alongside ownership, release plans, and critical workflows when setting remediation priorities.

For a realized defect, the relevant factor is the consequence that occurred. For a risk exposure, it is the credible potential consequence and the reachability of the harmful path. For an inefficiency, it is avoidable work rather than total resource use. Table~\ref{tab:severity-factors} gives the consequence and waste values. The evidence must support the selected factor: task failure does not establish financial loss, and extra calls do not establish waste when they provide necessary verification or recovery.

\begin{table*}[t]
\centering
\small
\renewcommand{\arraystretch}{1.12}
\begin{tabularx}{\textwidth}{@{}p{0.42in}XX@{}}
\toprule
Points & Consequence for a defect or credible exposure & Avoidable waste for an inefficiency\\
\midrule
10 & Small, localized adverse effect. & Small avoidable work bounded to one operation.\\
25 & Material but reversible degradation. & Meaningful but bounded avoidable work.\\
40 & Task failure or a materially wrong result. & Substantial or repeated avoidable work.\\
65 & Major supported harm in a high-consequence domain. & Waste dominates cost or latency, or persists across stages or turns.\\
85 & Severe, permanent, or irreversible harm. & Work continues until intervention or a hard limit and threatens viability.\\
\bottomrule
\end{tabularx}
\caption{Severity factor values. Consequence and waste have separate definitions even when their point values coincide. Potential consequence applies only to an evidenced reachable path.}
\label{tab:severity-factors}
\end{table*}

\begin{table*}[t]
\centering
\small
\renewcommand{\arraystretch}{1.12}
\begin{tabularx}{\textwidth}{@{}p{0.76in}p{1.38in}Xp{0.42in}@{}}
\toprule
Scope band & Distinct affected sessions & Affected share (20+ analyzed sessions) & Points\\
\midrule
Isolated & 1 & Less than 1\% & 0\\
Limited & 2--3 & 1\% to less than 10\% & 3\\
Repeated & 4--10 & 10\% to less than 30\% & 7\\
Broad & 11--100 & 30\% to less than 60\% & 12\\
Widespread & More than 100 & At least 60\% & 15\\
\bottomrule
\end{tabularx}
\caption{Affected-scope points. Use the higher band from the count and, when at least 20 sessions were analyzed, the percentage columns.}
\label{tab:severity-scope}
\end{table*}

Let $C$ denote observed consequence, $C_p$ potential consequence, $W$ avoidable waste, and $A$ affected-scope points. For risk, the reachability multiplier $R$ is 0.4 for privileged access with several uncommon preconditions, 0.6 for authentication or active user interaction, 0.8 for ordinary application behavior, and 1.0 for a public or reliably automatable path. The scores are
\begin{align}
 s_d &= \min(100, C+A),\\
 s_r &= \min(100, \lceil C_pR \rceil+A),\\
 s_i &= \min(100, W+A).
\end{align}
The ceiling rounds the adjusted potential consequence up to the next integer. An unreachable hypothetical path does not establish a risk occurrence; when harm has occurred, that occurrence is assessed as a realized defect.

Affected scope counts distinct affected sessions for each nature. Repeated spans or observations in one session do not increase the count. Table~\ref{tab:severity-scope} gives the count and percentage bands. The percentage applies only when at least 20 sessions were analyzed, and the higher of the count and percentage bands determines $A$. Its denominator is the set actually analyzed for that assessment. The calculation records the counts and population alongside the score so that later assessments can be interpreted against their evidence.

A Signal has a score for each nature represented by its occurrences. Its overall severity is $S=\max(s_d,s_r,s_i)$, using only those represented natures. Summing would mix different meanings and depend on the number of natures recorded; averaging could hide a rare severe defect. The display levels are Low for 1--39, High for 40--79, and Critical for 80--100. The largest consequence or waste value is 85 points and the largest scope value is 15, so the current tables cannot exceed 100; the explicit cap guards against later table changes.

The model assesses the semantic factors from the evidence. Code derives scope, looks up the fixed values, performs the arithmetic, and determines the display level. The explanation identifies the evidence for each factor and the nature that determines the overall score.

For illustration, a task failure affecting five of 30 analyzed sessions contributes 40 consequence points. Both the count of five and the 16.7\% affected share select the repeated band, adding seven scope points. The score is $40+7=47$, with a High display level. This is an arithmetic example, separate from the case in Section~\ref{sec:case}. New evidence can change scope or establish a different consequence, but a higher count alone does not justify a more severe consequence category.

\subsection{Suggesting remediations}
A remediation specifies an action a developer can take and how to check its effect. Depending on the evidence, the action may concern instructions, context, tool integration, or code. We kept recommendations specific to the available evidence: a tool contract and invalid invocation can establish how an argument should change, but locating the exact code requires implementation context.

The action may prevent the problem or help detect it. Correcting argument construction can prevent an invalid call; adding an evaluator can identify similar calls later. The recommendation needs to make that intended effect clear. Validation then checks that the change addresses the problem while preserving correct behavior. Fewer calls or lower cost count as an improvement only if the agent still completes the task correctly. Signals supplies the diagnosis and concrete actions; developers implement and validate the changes.

\subsection{Insurance-intake example}\label{sec:case}
In an audited synthetic insurance-intake task, a false conflict led to human review instead of case registration. The agent read a submission and supporting document, compared their policy references, and decided whether to register a case or send it for human review. The submission contained \code{POL123}; extraction returned \code{Policy POL123}. The second value added a field label to the same identifier. The workflow nevertheless reported a conflict, routed the request to human review, and did not register a case in that attempt.

\begin{table}[htbp]
\centering
\small
\setlength{\tabcolsep}{3pt}
\renewcommand{\arraystretch}{1.14}
\begin{tabularx}{\columnwidth}{@{}p{0.94in}X@{}}
\toprule
Taxonomy problem & Evidence supporting the diagnosis\\
\midrule
Factual or logical error & \code{POL123} and \code{Policy POL123} identify the same policy. The added field label did not support the reported \code{policy\_reference} conflict.\\
\bottomrule
\end{tabularx}
\caption{The taxonomy diagnosis in the synthetic insurance-intake case. Generic policy identifiers replace the source values.}
\label{tab:case}
\end{table}

Signals reported ``Unsupported conflict flags'' and ``False conflict detection''. Table~\ref{tab:case} connects these findings to the taxonomy and their evidence. The underlying \emph{Factual or logical error} was the comparison of equivalent identifiers as different. Human-review routing and missing registration showed its effect; they did not independently establish additional defects or financial harm. Human review remains appropriate for a genuine conflict.

We proposed comparing normalized policy identifiers while preserving real differences. Checks should cover equivalent forms (\code{POL123} and \code{Policy POL123}), different references (\code{POL123} and \code{POL999}), and missing references. Different or missing values should still follow the applicable review policy. The audit did not include an implemented repair. Appendix~\ref{app:payment} gives a simpler payment-tool illustration.

\FloatBarrier

\section{Taxonomy for Agent Diagnostics}\label{sec:taxonomy}
We built the taxonomy to give the algorithm a consistent definition of the problems it should find. A task outcome alone is insufficient: two failed tasks can require different fixes, and a completed task can still contain a risk exposure or unnecessary work. The catalog defines 45 problem types across eight families, summarized in Table~\ref{tab:taxonomy}. Appendix~\ref{app:taxonomy} gives every definition. The families organize the catalog; they do not establish a problem's root cause. Shared definitions also give reviewers a basis for comparing findings across tasks, agent designs, and diagnostic methods.

\begin{table*}[t]
\centering
\small
\renewcommand{\arraystretch}{1.15}
\begin{tabularx}{\textwidth}{@{}p{2.05in}p{0.35in}X@{}}
\toprule
Diagnostic family & Types & Diagnostic focus\\
\midrule
Instruction and specification & 5 & Following the accepted objective, explicit requirements, and permitted scope.\\
Knowledge, context, and retrieval & 6 & Obtaining, retaining, and citing the evidence needed for the task.\\
Reasoning, decision, and output & 8 & Producing supported, correct, complete results and truthful completion claims.\\
Tool, action, and integration & 8 & Choosing and calling tools, interpreting their results, and respecting intended effects.\\
Orchestration, coordination, and state & 5 & Routing work, transferring information and state, making progress, and stopping correctly.\\
Runtime and platform reliability & 3 & Meeting execution requirements despite provider, capacity, authentication, or configuration failures.\\
Safety, security, privacy, and policy & 6 & Respecting trust, authorization, confidentiality, and safe output boundaries.\\
Efficiency and resource use & 4 & Avoiding unnecessary actions, retries, retrievals, and context while preserving acceptable outcomes.\\
\bottomrule
\end{tabularx}
\caption{The 45 problem types and their eight diagnostic families.}
\label{tab:taxonomy}
\end{table*}

\subsection{How the taxonomy was built}
To decide which behaviors the catalog needed to distinguish, we reviewed research on instruction following and priority~\cite{r7,r8,r9,r10}; agent behavior, coordination, and stateful tool use~\cite{r2,r11,r3,r12,r13,r1,r14,r15}; retrieval, answer support, and citations~\cite{r16,r4,r17,r18,r19,r20}; and safety, security, refusal, and bias~\cite{r21,r5,r22,r23,r24,r25,r26,r27}. We also used telemetry and reliability guidance to identify runtime evidence~\cite{r28,r29,r30,r31,r32}, and cost--quality studies to inform our treatment of avoidable resource use~\cite{r33,r34}.

We consolidated these sources by asking what evidence would establish each problem and what action would address it. For each type, we recorded a definition, required evidence, examples, exclusions, and boundaries with neighboring types. We separated a proposed cause from the observed behavior and its consequence. We split similar symptoms when their evidence requirements differed, and merged overlapping descriptions when they described the same observable violation. We did not adopt benchmark categories unchanged.

The resulting catalog includes runtime and platform failures, safety boundaries, and avoidable resource use alongside failures in the agent's decisions. Each family therefore requires evidence appropriate to the problem: a provider outage, an unauthorized action, and a redundant retrieval cannot be established in the same way. Source frequency does not establish production prevalence. We disclosed the public benchmark families used in this research and did not treat them as unseen evidence. The catalog can be extended as new agent architectures, tools, and use cases produce problems that the current definitions do not capture.

\subsection{Evidence and diagnostic boundaries}
We defined types around observable requirements so that a diagnosis could be checked against the execution. Invalid tool arguments concern the invocation; an invalid tool result concerns the tool's response; and ignoring a valid result concerns the agent's subsequent decision. Similar task outcomes can therefore require different evidence and remediation actions.

A diagnosis must establish both the applicable requirement and the behavior that violated it. For example, invalid arguments require the tool contract and the submitted arguments. A false success claim requires an explicit completion claim contradicted by execution records or authoritative state. Missing telemetry alone does not prove that an action never happened, and private or missing reasoning cannot establish a mismatch between a stated decision and an action.

Problem type remains separate from the outcome distinction used in Section~\ref{sec:method}. A type can describe a realized defect, a supported risk exposure, or an inefficiency when the evidence warrants it. Attack-like text alone does not establish a security failure, and repeated work may be justified by verification, recovery, pagination, or freshness checks. We also distinguished no supported problem from insufficient evidence and from a candidate problem outside the catalog. Several descriptions of one violation do not make it several events; independently supported problems can coexist in the same execution.

\FloatBarrier

\section{Dataset}\label{sec:dataset}
We assembled a reference dataset from recorded agent executions in public benchmarks. The sources were AgentProcessBench~\cite{r35}, SWE-rebench / OpenHands~\cite{r36}, TraceElephant~\cite{r37}, MAD / MAST~\cite{r2}, the original $\tau$-bench~\cite{r1}, and recorded SWE-bench trajectories~\cite{r6}. We also included AgentDojo~\cite{r5}, Who\&When~\cite{r15}, TRAIL~\cite{r38}, RootSE~\cite{r39}, and AgentRx~\cite{r3}. Together, these sources covered coding, tool use, coordination, and security evaluation.

For each session, the detector received the available task, instructions, tool information, actions, and observations. We excluded native benchmark outcomes and reference labels from its inputs. This kept the reference labels separate from the execution evidence used for detection.

LLM judges created and reviewed the reference labels under human oversight. We organized the resulting reference occurrences into reference Signals within evaluation batches so that we could evaluate individual findings and the issues formed from them. Each reference occurrence belonged to exactly one reference Signal; the same issue could appear in several batches.

\section{Experiments and Results}\label{sec:experiments}
To test how accurately our algorithm identified errors and inefficiencies, we evaluated its findings against the reference dataset in Section~\ref{sec:dataset}. We first measured agreement on individual occurrences and on the Signals formed from them. We then compared the full workflow with a one-pass control to test whether the workflow added value beyond supplying the taxonomy to the model. Separate session-level checks examined detection and agreement across repeated runs.

\subsection{Evaluation setup}
The Signals detector used GPT-6.1 Sol. For all occurrence results, we used the same frozen reference dataset and fixed predictions. Code matched predictions to references and calculated the metrics; we did not call a model judge for either step.

We allowed a predicted occurrence to match a reference occurrence only when they belonged to the same session, had the same known component type and normalized component name, and shared at least one problem-type ID across their three ranked candidates. Name normalization accounted for case, camel-case boundaries, Unicode form, and separators. We then computed a deterministic maximum one-to-one matching, giving each prediction and reference occurrence credit at most once. Unmatched predictions and references remained in the metric denominators. Descriptions, evidence-span overlap, severity, and remediation did not enter the matcher.

At Signal level, we used two checks. First, we measured reference coverage: the fraction of reference Signals with at least one matched occurrence, and the fraction with at least half their occurrences matched. These matches could be spread across several predicted Signals. Second, we matched predicted and reference Signals within the same evaluation batch, requiring the same component type, normalized component name, and at least one shared problem-type ID. A maximum one-to-one matching again gave each Signal credit at most once. This comparison used the component and problem-type tags, without using occurrence IDs, counts, or membership overlap.

These rules measured agreement on the component and problem type. They did not verify the wording of a diagnosis, the cited evidence, severity calibration, or the correctness of a proposed fix. The matching rule could also connect different events on the same component, including when only a lower-ranked problem-type alternative matched.

\subsection{Detection accuracy}
Occurrence matching gave \textbf{90.01\% precision, 90.98\% recall, and 90.49\% F1}. An unmatched prediction could reflect a component or type disagreement, or an extra repeated report; it was not automatically a factually false description.

The matched occurrences covered \textbf{92.86\%} of reference Signals with at least one match and \textbf{92.16\%} with at least half their occurrences matched. The separate Signal comparison gave \textbf{92.90\% precision, 90.01\% recall, and 91.43\% F1}. Coverage depended on occurrence matching, while this second comparison required agreement on the component and problem-type tags without an occurrence-overlap requirement.

We also measured detection at the session level, obtaining \textbf{97.26\% precision and 97.47\% recall}. This check required only a prediction somewhere in the run, rather than recovery of the correct individual problems.

\subsection{Comparison with direct Signal generation}
To separate the effect of the workflow from that of the model and taxonomy, we compared Signals with a one-pass direct-generation control. The control used the same model and complete 45-type taxonomy and generated complete Signals directly from batches of complete sessions. We scored both systems against the same reference Signals, using the same evaluation cohort and deterministic Signal matcher. Table~\ref{tab:comparison} reports the comparison.

\begin{table}[htbp]
\centering
\small
\setlength{\tabcolsep}{3.5pt}
\renewcommand{\arraystretch}{1.15}
\begin{tabularx}{\columnwidth}{@{}Xrrr@{}}
\toprule
System & Precision & Recall & F1\\
\midrule
One-pass direct generation & 71.11\% & 76.46\% & 73.69\%\\
Signals pipeline & 92.90\% & 90.01\% & 91.43\%\\
\bottomrule
\end{tabularx}
\caption{Signal matching with the same model, taxonomy, evaluation cohort, and deterministic matcher.}
\label{tab:comparison}
\end{table}

The Signals pipeline improved precision by \textbf{21.79 percentage points}, recall by \textbf{13.56 points}, and F1 by \textbf{17.75 points}. The larger precision improvement indicated that the workflow substantially reduced unmatched Signal predictions. In this comparison, the workflow contributed beyond giving the taxonomy to the model in a single generation step. The results came from selected benchmarks; correlated tasks and exposure to benchmark families during taxonomy development constrain their generalization.

\subsection{Robustness and recurrence}
In separate fixed robustness runs, we measured agreement on whether a session contained a finding across repeated runs. Agreement was \textbf{97.0--98.5\%}; removing some telemetry reduced it to \textbf{74.2\%}. These were binary session-presence comparisons, separate from the occurrence-matching scores above.

We also examined exploratory lifecycle histories involving returning issues and periods without recurrence. We did not implement and test the proposed remediations, so these histories did not establish whether a fix worked. The runs also did not establish production cost savings or a reduction in diagnosis time.

\section{Conclusion}\label{sec:conclusion}
We presented Signals as a way to connect a \textbf{45-type diagnostic taxonomy} to execution evidence, severity assessment, and \textbf{concrete remediation suggestions}. The taxonomy gave us definitions for the problems to identify; the diagnostic workflow connected individual findings to issues a developer could investigate and fix.

On the reference dataset, the Signals pipeline achieved \textbf{91.43\% Signal-matching F1}, compared with \textbf{73.69\%} for a one-pass control using the same model and taxonomy. This comparison showed that the workflow added value beyond direct taxonomy-guided generation. The scores measured agreement on the component and problem type. They did not establish the accuracy of severity assessments or the effectiveness of proposed remediations, which require separate validation.

\FloatBarrier

\appendix

\section{Diagnostic Taxonomy Catalog}\label{app:taxonomy}

This catalog defines the 45 problem types introduced in Section~\ref{sec:taxonomy}. Each diagnosis requires the applicable task, policy, or interface requirement and supporting execution or configuration evidence. The definitions describe observable behavior; they do not establish hidden causes or intent. Realized defects, risk exposures, and inefficiencies are distinguished as described in the main text.

\subsection{Instruction and specification}
\begin{itemize}

\item \textbf{Constraint or instruction-priority violation.} An action or output violates an applicable explicit constraint, or follows a lower-priority instruction over a conflicting higher-priority instruction. The governing instructions and violating behavior must both be available.

\item \textbf{User intent misinterpreted.} The agent adopts a materially different objective from the user's request before execution. The request, accepted objective or plan, and resulting behavior must show the mismatch.

\item \textbf{Required clarification not requested.} The agent acts or stops without asking for a required value that the user could supply and policy permits it to request. The missing value must explain an incorrect action or outcome.

\item \textbf{Inappropriate refusal.} The agent refuses an allowed, supported request despite sufficient information and no policy requiring refusal. Correct safety refusals, access denials, and user cancellations are excluded.

\item \textbf{Goal drift.} After correctly accepting the task, the agent pursues a materially different objective or expands its scope without authorization. The initial objective, later change, and resulting actions must be visible; authorized user revisions are excluded.

\end{itemize}

\subsection{Knowledge, context, and retrieval}
\begin{itemize}

\item \textbf{Required evidence not retrieved.} Retrieval omits source material required by the task even though it exists in the configured corpus and is retrievable. The corpus or relevance reference must establish what was missing.

\item \textbf{Irrelevant or noisy retrieval.} Retrieved material is irrelevant to the task and degrades the result or adds avoidable processing cost. A large retrieved context alone does not establish this problem.

\item \textbf{Stale or false retrieved context.} The retrieval path supplies content that is outdated or false for the request. A trusted reference or time and version evidence must establish the discrepancy; a risk exposure requires a proven reachable path.

\item \textbf{Required context or memory lost.} Information available earlier is absent, truncated, or replaced when the agent later needs it. Earlier records, the later context, and the resulting behavior must show the loss mattered.

\item \textbf{Context window limit exceeded.} The assembled input exceeds, or is proven able to exceed, the active model's context limit before required completion. The model version, limit, and counted input must establish the condition.

\item \begingroup\widowpenalty=10000
\textbf{Citation source mismatch or unsupported citation.} A cited source does not support its paired claim, or a required citation is missing. Assessment uses the claim, source content, and applicable citation \mbox{requirement}.\par\endgroup

\end{itemize}

\subsection{Reasoning, decision, and output}
\begin{itemize}

\item \textbf{Unsupported or invented claim.} A material claim is not supported by the evidence available to the agent. Establishing that the claim is false is a separate question; incomplete records alone cannot establish lack of support.

\item \textbf{Factual or logical error.} An output or decision contradicts established facts, a deterministic calculation, or a governing logical condition. The diagnosis requires a reference that establishes the contradiction.

\item \textbf{Incomplete or prematurely terminated result.} The agent returns or stops before satisfying a required part of the accepted task, including a required final answer. Correctly scoped partial results, valid refusals, and user cancellations are excluded.

\item \textbf{Verification absent or incorrect.} A required check is skipped, incomplete, or reports a pass inconsistent with a directly checkable condition. Optional review and checks that correctly block bad output are excluded.

\item \textbf{Output schema or format violation.} An agent output violates the syntax, fields, or format required by its consumer. The output contract and raw output must establish the violation; tool arguments are a separate type.

\item \textbf{Action and stated decision mismatch.} An executed action or delivered artifact contradicts the same component's visible committed decision while the objective remains unchanged. Authorized plan revisions and missing private reasoning are excluded; a more specific tool or routing violation uses that type.

\item \textbf{False success claim.} The agent claims an action or task succeeded, but execution records or authoritative state establish that it did not. Missing records alone do not prove non-completion, and the diagnosis does not infer deceptive intent.

\item \textbf{Success check subverted.} An observed modification or check-specific shortcut makes a success check pass while an independently established task requirement remains unmet. Authorized test corrections and legitimate solutions are excluded; intent is not inferred.

\end{itemize}

\subsection{Tool, action, and integration}
\begin{itemize}

\item \textbf{Wrong tool selected.} The agent invokes an unsuitable or impermissible tool when a suitable available tool or correct path without a tool exists. The task, available tools, and selection requirement must establish the error.

\item \textbf{Required tool call omitted.} A required tool call is absent before the agent acts, answers, or stops, without selecting an impermissible tool instead. The call must be required under the accepted task and current state.

\item \textbf{Invalid tool arguments.} A tool invocation violates its schema, required values, types, or domain preconditions. The tool contract, submitted arguments, and validation or rejection establish the problem.

\item \textbf{Tool interface contract mismatch.} The advertised tool interface disagrees with the interface actually exposed across an integration boundary. Versioned definitions must establish the mismatch and a resulting failure or reachable invocation path.

\item \textbf{Tool result invalid.} A tool reports success but produces a result that violates its structural or semantic contract. A reported failure is a different event; a risk exposure requires a proven reachable invalid result.

\item \textbf{Tool result misinterpreted or ignored.} The agent receives a valid tool result but its next decision contradicts or fails to use material information from it. The result and subsequent behavior must establish the error.

\item \textbf{Tool execution error not handled.} A tool error or invalid result is followed by a separate failure to recover, validate, degrade safely, or report as required. The error alone does not establish this additional problem.

\item \textbf{Unintended tool side effect.} A tool changes state beyond or differently from its authorized intended effect. The intended action, tool execution, and state before and after the call must establish the unexpected change.

\end{itemize}

\subsection{Orchestration, coordination, and state}
\begin{itemize}

\item \textbf{Invalid route, sequence, or decomposition.} The workflow chooses an incorrect node, order, branch, or task decomposition for the accepted objective. The expected workflow and actual execution must show that the choice changed execution or outcome.

\item \textbf{Handoff information lost or ignored.} Information required at a handoff is omitted, withheld, or received and ignored by the destination agent or node. Source state, transferred information, and downstream behavior must show the failure.

\item \textbf{Workflow state not propagated or reset.} A workflow transition fails to preserve or update shared state needed by a later node. State snapshots and the expected transition must establish the missing, stale, or reset value.

\item \textbf{Workflow loop or deadlock.} The workflow repeats a control state without useful progress or waits cyclically, preventing it from reaching completion. The repeated states or wait cycle and progress evidence must establish the problem.

\item \textbf{Termination condition missed.} Execution continues without an authorized purpose after an applicable stopping condition has been reached. Necessary reporting, cleanup, verification, and recovery are excluded; a transient error alone is not a stopping condition.

\end{itemize}

\subsection{Runtime and platform reliability}
\begin{itemize}

\item \textbf{Provider or runtime timeout/unavailable.} A model provider or workflow runtime exceeds its deadline or is unavailable, preventing execution from meeting its service requirement. A correctly handled timeout without a service impact is excluded.

\item \textbf{Rate or resource exhaustion.} Quota, capacity, memory, concurrency, or another resource limit is exhausted or proven reachable under expected load, impairing the agent execution. Metrics, errors, or configuration and load evidence must establish the condition.

\item \textbf{Runtime authentication or configuration failure.} The model provider or runtime cannot execute because required credentials, endpoints, deployments, or settings are invalid or missing. A risk exposure requires configuration evidence of a reachable failure before an outage.

\end{itemize}

\subsection{Safety, security, privacy, and policy}
\begin{itemize}

\item \textbf{Untrusted-instruction boundary failure.} Untrusted content controls behavior across a required instruction boundary, or configuration proves a reachable takeover path under a stated threat model. Attack-like text alone does not establish the problem.

\item \textbf{Sensitive information disclosure.} Sensitive information reaches an unauthorized destination or is unnecessarily placed in a reachable unsafe context. Data classification, authorization, and the actual or reachable information flow must establish the problem.

\item \textbf{Excessive or unauthorized agency.} An agent takes or can take actions beyond the approved, authorized, or confirmation-bounded scope. Excessive permissions without an observed harmful action constitute a risk exposure when the reachable path is established.

\item \textbf{Unsafe or disallowed content produced.} The agent produces content prohibited by the application's explicit safety policy in the evaluated context. The policy and output must establish the violation; an ordinary factual error alone is insufficient.

\item \textbf{Policy or access-control bypass.} A required authorization, policy, guardrail, or trust check is bypassed, or configuration establishes a reachable bypass. The applicable control, permissions, and action or path must support the diagnosis.

\item \textbf{Insecure model/tool output handling.} Untrusted model or tool output reaches an interpreter, renderer, command, or privileged destination without required validation, encoding, or isolation. Evidence must establish an executed effect or reachable dangerous path.

\end{itemize}

\subsection{Efficiency and resource use}
\begin{itemize}

\item \textbf{Repeated or duplicate action.} A successful, safe task repeats an equivalent completed action without adding needed evidence, state, safety, or recovery value. Required verification, pagination, and other justified repetition are excluded.

\item \textbf{Retry thrash.} A task succeeds or safely degrades but repeatedly retries a failing operation beyond a defensible retry policy without useful adaptation. Error history, timing, policy, and the acceptable outcome must establish avoidable work.

\item \textbf{Redundant retrieval.} A successful, safe task retrieves equivalent evidence already available without a justified freshness, coverage, or verification need. The retrieval sequence and available context must show no material evidence gain.

\item \textbf{Avoidable context bloat.} A successful, safe task processes irrelevant or duplicate context that causes avoidable cost or latency. A smaller context that preserves quality, or equivalent evidence, must establish the waste; size alone is insufficient.

\end{itemize}

\section{Payment Tool Illustration}\label{app:payment}
We considered two synthetic sessions requesting a test payment of 100 cents. The supplied \code{create\_payment} contract requires \code{amount\_cents}. One agent instead sent \code{amount}; the tool rejected the call and the task failed. A control session used the required field and succeeded. Table~\ref{tab:payment} shows the records.

\par\medskip\noindent
\begin{minipage}{\columnwidth}
\centering
\small
\setlength{\tabcolsep}{3pt}
\renewcommand{\arraystretch}{1.12}
\begin{tabularx}{\columnwidth}{@{}p{0.70in}XX@{}}
\toprule
Record & Rejected session & Control session\\
\midrule
Request & Payment of 100 cents & Payment of 100 cents\\
Arguments & amount: 100 & amount\_cents: 100\\
Tool result & Required field missing & Payment created\\
Task result & Failed & Succeeded\\
\bottomrule
\end{tabularx}
\captionof{table}{Synthetic executions using the same payment-tool contract.}
\label{tab:payment}
\end{minipage}
\par\medskip

The problem type was \emph{Invalid tool arguments}, with a realized defect. The contract, arguments, and rejection response established the error. The remediation was to use \code{amount\_cents} while preserving the requested amount. Checks should cover a valid payment, a missing amount, and an invalid value, including verification that rejected requests create no payment. These sessions were hand-authored illustrations, not measured detection or repair results.

\end{document}